\documentclass[sigconf]{acmart}

\usepackage{booktabs} 
\usepackage{subcaption}
\usepackage{multirow}
\usepackage{tabu}
\usepackage{graphicx}
\PassOptionsToPackage{table,xcdraw}{xcolor}
\usepackage{colortbl}
\usepackage{enumitem}

\newcommand{\commenta}[1]

\setcopyright{rightsretained}

\begin{document}
\title{Building Adversary-Resistant Deep Neural Networks without Security through Obscurity}

\author[Q. Wang]{Qinglong Wang}
\affiliation{McGill University}
\affiliation{Pennsylvania State University}

\author[W. Guo]{Wenbo Guo}\authornote{The first and second author contribute equally.}
\affiliation{Pennsylvania State University}
\author[K. Zhang]{Kaixuan Zhang}
\affiliation{Pennsylvania State University}
\author[A. Ororbia II]{Alexander G. Ororbia II}
\affiliation{Pennsylvania State University}
\author[X. Xing]{Xinyu Xing}
\affiliation{Pennsylvania State University}
\author[L. Lin II]{Lynn Lin}
\affiliation{Pennsylvania State University}
\author[X. Liu II]{Xue Liu}
\affiliation{McGill University}
\author{C. Lee Giles}
\affiliation{Pennsylvania State University}

\begin{abstract}

Deep neural networks (DNNs) have proven to be quite effective in a vast array of machine learning tasks, with recent examples in cyber security and autonomous vehicles. Despite the superior performance of DNNs in these applications, it has been recently shown that these models are susceptible to a particular type of attack that exploits a fundamental flaw in their design. This attack consists of generating particular synthetic examples referred to as adversarial samples. These samples are constructed by slightly manipulating real data-points in order to ``\emph{fool}'' the original DNN model, forcing it to mis-classify previously correctly classified samples with high confidence. Addressing this flaw in the model is essential if DNNs are to be used in critical applications such as those in cyber security. 

Previous work has provided various learning algorithms to enhance the robustness of DNN models, and they all fall into the tactic of ``security through obscurity''. This means security can be guaranteed only if one can obscure the learning algorithms from adversaries. Once the learning technique is disclosed, DNNs protected by these defense mechanisms are still susceptible to adversarial samples. In this work, we investigate this issue shared across previous research work and propose a generic approach to escalate a DNN's resistance to adversarial samples. More specifically, our approach integrates a data transformation module with a DNN, making it robust even if we reveal the underlying learning algorithm. To demonstrate the generality of our proposed approach and its potential for handling cyber security applications, we evaluate our method and several other existing solutions on datasets publicly available, such as a large scale malware dataset, MNIST and IMDB datasets. Our results indicate that our approach typically provides superior classification performance and resistance in comparison with state-of-art solutions.    

\end{abstract}

\maketitle

\section{Introduction}
\label{sec:intro}


Beyond highly publicized victories in automatic game-playing as in
Go~\cite{silver2016mastering}, there have been many successful applications of
deep neural networks (DNN) in image and speech recognition. Recent explorations
and applications include those in medical imaging~\cite{bar2015deep, xu2014deep}
and self-driving cars~\cite{hadsell2009learning, farabet2012scene}. In the
domain of cybersecurity, security companies have demonstrated that deep learning
could offer a far better way to classify all types of
malware~\cite{pascanu2015malware, dahl2013large, yuan2014droid}.


Despite its potential, deep neural networks (DNN), like all other machine learning
approaches, are vulnerable to what is known as adversarial
samples~\cite{huang2011adversarial, barreno2006can}. This means that they can be
easily deceived by non-obvious and potentially dangerous
manipulation~\cite{42503, nguyen2015deep}. To be more specific, an attacker
could use the same training algorithm, back-propagation of errors, and a
surrogate dataset to construct an auxiliary model. Since this model could
provide the attacker with a capability of exploring a DNN's blind spots,  he
can, with minimal effort, craft an {\em adversarial sample} -- a synthetic
example generated by slightly modifying a real example in order to make the deep
learning system ``believe'' the sample subtly perturbed belongs to an incorrect class
with high confidence.


According to a recent study~\cite{Goodfellow14}, adversarial samples occur in a subspace
relatively broad, which means it is impractical to build a defense that can rule out
all adversarial samples. As such, the design principle followed by existing
defense mechanisms is not to harden a DNN model naturally resistant to any
adversarial samples. Rather, they focus on hiding that subspace, making
adversaries difficult in finding impactful adversarial samples. For example,
representative defenses -- {\em adversarial training}~\cite{Goodfellow14} and {\em defensive
distillation}~\cite{papernot2015distillation} both increase the complexity of original DNNs with the
goal of making adversarial samples -- impactful for original DNNs -- no longer
effective.


However, in this work, we show that the defenses proposed are far from ideal and
even considered a dangerous practice. In particular, we demonstrate existing
defense mechanisms all follow the approach of ``security through obscurity'', in
which security is achieved by keeping defenses obscured from adversaries.
Frankly speaking, defenses following this approach indeed mitigate the
adversarial sample problem. However, when applied to security critical
applications such as malware classification, they become particularly
disconcerting.

In the past, there have been a huge amount of debates on security through
obscurity, and a general consensus has been reach. That is, obscurity is a
perfectly valid security tactic but it cannot be trusted for security. Once
design or implementation is uncovered, users totally lost all the security
gained by obscurity. To regain the security through obscurity, one has to come
up with a completely new design or implementation. As such, Kerckhoffs'
principle~\cite{Kerckhoffs1883} suggests obscurity can be used as a layer of defense, but it
should never be used as the only layer of defense.


Inspired by this, we propose a new mechanism to escalate a DNN's resistance to
adversarial samples. Different from existing defenses, our proposed approach
unnecessitates model obscurity. In other words, even though we reveal the model,
it will still be more than burdensome for adversaries to craft adversarial
samples.

More specifically, we arm a standard DNN with a data transformation 
module, which projects original data 
input into a new representation before they are passed through the
consecutive DNN. This can be used as a defense for the following two reasons.
First, data transformation can potentially stash away the space of
adversarial manipulations to a carefully designed hyperspace. This makes
attackers difficult in finding adversarial samples impactful for the armed DNN.
Second, as we will theoretically prove in Section~\ref{sec:framework}, a data transformation 
module carefully designed can exponentially increase computation complexity for an attacker
to craft impactful adversarial samples. This means that, even though an attacker
compromises obscurity and has the full knowledge about the armed DNN model (i.e., the 
training algorithm, dataset and hyper-parameters), he
still cannot launch the attack -- detrimental to DNNs enhanced by other
adversary-resistant techniques -- nor jeopardize model resistance.


The approach proposed in this work is beneficial for the following reasons.
First, it escalates a DNN's resistance to adversarial samples with better
security assurance. Second, our approach ensures that a DNN maintains
desirable classification performance while requiring only minimal modification
to existing architectures. Third, while this work is primarily
motivated by the need to safeguard DNN models used in critical security
applications, it should be noted that the proposed technique is general and can
be adopted to other applications where deep learning is popularly applied, such
as image recognition and sentiment analysis. We demonstrate this applicability 
using publicly-available datasets in Section~\ref{sec:eval}.


In summary, this work makes the following contributions.

\begin{itemize}
	
    \item We propose a generic approach to facilitate the development of
    adversary-resistant DNNs without following the tactic of security
    through obscurity.

	\item Using our approach, we develop an adversary-resistant DNN, and
	theoretically prove its resistance cannot be jeopardized even if the model
	is fully disclosed.
	
    \item We evaluate the classification performance and robustness of our
	adversary-resistant DNN and compare it with that of existing defense 
	mechanisms. Our result shows that our DNN exhibits similar -- sometimes even better -- classification 
	performance but with superior model resistance.

\end{itemize}

The rest of this paper is organized as follows. Section~\ref{sec:background}
introduces the background of DNNs and adversarial sample problem.
Section~\ref{sec:problem} discusses existing defense mechanisms and defines the problem scope of our research. Section~\ref{sec:framework} presents our generic approach. In Section~\ref{sec:eval}, we develop and evaluate DNNs in the context of
image recognition, sentiment analysis and malware classification. Finally, we
conclude this work in Section~\ref{sec:conclusion}.


\begin{figure}[t]
	\begin{center}
		\includegraphics[width=0.45\textwidth]{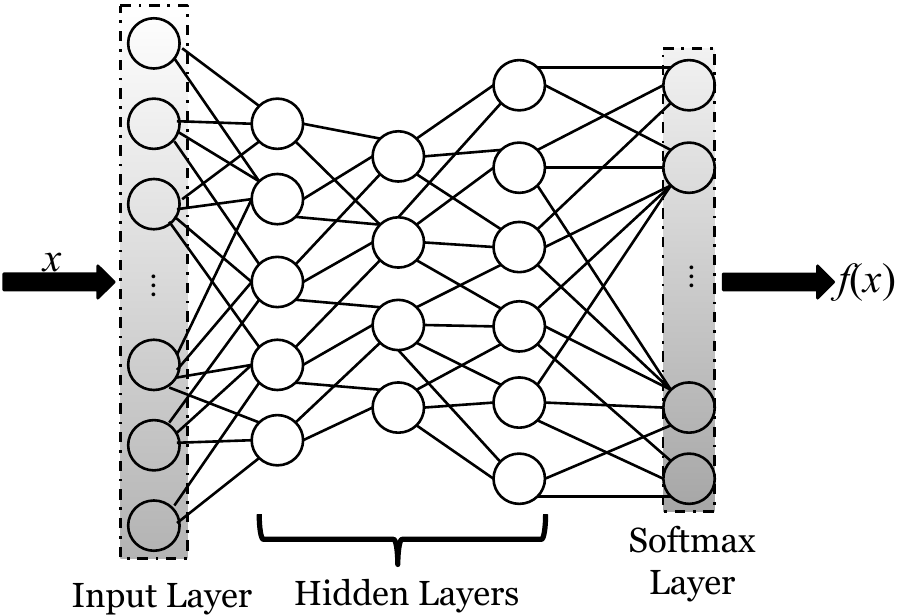}
		\caption{A feed-forward deep neural network with three hidden layers and a softmax output layer.}
		\label{fig:dnn}
	\end{center}
\end{figure}

\section{Background}
\label{sec:background}

A typical DNN architecture consists of multiple successive layers of processing elements, or so-called ``neurons''. Each processing layer can be viewed as learning a different, more abstract representation of the original multidimensional input distribution. As a whole, a DNN can be viewed as a highly complex function, $f(\cdot)$ that is capable of nonlinearly mapping an original high-dimensional data point to a lower dimensional output. In  this  section, we briefly introduce  the  well-established DNN  model, followed by the description of the adversarial learning problem.

\subsection{Deep Neural Networks}

As is graphically depicted in Figure~\ref{fig:dnn}, a DNN contains an input layer, multiple hidden layers, and an output layer. The input layer takes in each data sample in the form of a multidimensional vector. Starting from the input, computing the pre-activations of each subsequent layer simply requires, at minimum, a matrix multiplication (where a weight/parameter vector, with length equal to the number of hidden units in the target layer, is assigned to each unit of the layer below) usually followed by summation with a bias vector. This process roughly models the process of a layer of neurons integrating the information received from the layer below (i.e., computing a pre-activation) before applying an elementwise activation function\footnote{There are many types of activations to choose from, including the hyperbolic tangent, the logistic sigmoid, or the linear rectified function, etc.~\cite{Goodfellow-et-al-2016-Book}}. This integrate-then-fire process is repeated subsequently for each layer until the last layer is reached. The last layer, or output, is generally interpreted as the model's predictions for some given input data, and is often designed to compute a parameterized likelihood distribution using the softmax function (also known as multi-class regression or minimizing cross entropy). This bottom-up propagation of information is also referred to as \emph{feed-forward} inference~\cite{hinton2007learning}.

During the learning phase of the model, the DNN's predictions are evaluated by comparing them with known target labels associated with the training samples (also known as the``ground truth''). Specifically, both predictions and labels are taken as the input to $L(\cdot)$, a selected cost function. The DNN's parameters are then optimized with respect to this cost function using the method of steepest gradient descent, minimizing prediction errors on the training set. 

More formally, given $(X,Y)$: $\{(x_{1}, y_{1}), \dots, (x_{n}, y_{n})\}$, where $x_{i}\in\mathbb{R}^{m}$ is a data sample and $y_{i} \in \mathbb{R}^{k}$ is the corresponding data label, where, if categorical, is typically represented through a 1-of-\emph{k} encoding, the goal of model learning is to minimize the cost function represented by $\displaystyle\sum_{i=1}^{n} L(f(w; x_i); y_i)$, where $f(w; x_i)$ denotes the prediction of training sample $x_i$ and $w$ represents the weights and bias associated with the connections between neurons.

\begin{figure}[t]
	\begin{center}
		\includegraphics[width=0.45\textwidth]{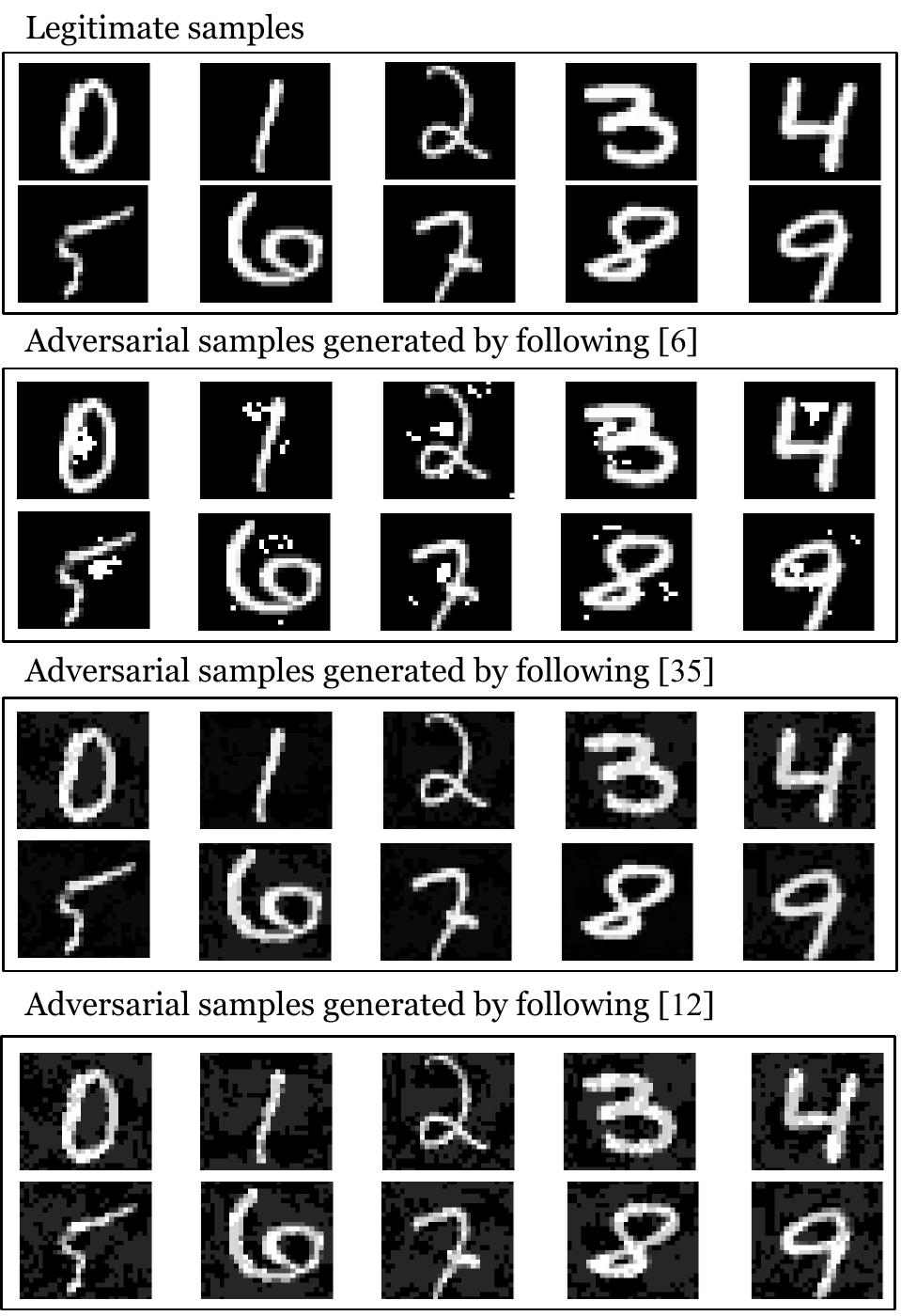}
		\caption{Legitimate samples and their corresponding adversarial samples generated by following various approaches. The adversarial samples are crafted by introducing to the legitimate samples pertubations nearly indistinguishable to the human eyes.}
		\label{fig:advsample}
	\end{center}
\end{figure}

\subsection{Adversarial Sample Problem}
\label{sec:advlearn}

An adversarial sample is a synthetic data sample crafted by introducing slight perturbations to a legitimate input sample (see Figure~\ref{fig:advsample}). In multi-class classification tasks, such adversarial samples can cause a DNN to classify themselves into a random class other than the correct one (sometimes not even a reasonable alternative). Recent research~\cite{Carlinioakland2017} demonstrates that, attackers can uncover such data samples through various approaches (e.g.,~\cite{42503, Goodfellow14, papernot2016limitations, sabour2015adversarial, bastani2016measuring, liu2016delving}) which can all be described as solving either optimization problem
\begin{equation}
  \begin{aligned}
  \label{eq:adv1}
	 &\hat{x} =\, arg\, \underset{\hat{x}}{\mathrm{max}}\,L(f(\hat{x}; w); y)\\
	              & \mathrm{s.t.} \hspace{0.2cm}\left \| \hat{x}-x \right \|_{p}<\varepsilon,
   \end{aligned}
\end{equation}
or optimization problem
\begin{equation}
  \begin{aligned}
  \label{eq:adv2}
	 &\hat{x} =\, arg\, \underset{\hat{x}}{\mathrm{min}}\,L(f(\hat{x}; w); \hat{y})\\
	              & \mathrm{s.t.} \hspace{0.2cm}\left \| \hat{x}-x \right \|_{p}<\varepsilon \\
	              & \hspace{0.7cm} \hat{y} \neq y.
   \end{aligned}
\end{equation}

Here, optimization problem~\eqref{eq:adv1} indicates that an attacker searches for adversarial sample $\hat{x}$, the prediction of which is as far as its true label, whereas optimization problem~\eqref{eq:adv2} indicates an attacker searches for adversarial sample $\hat{x}$ so that its prediction is as close as target label $\hat{y}$ where $\hat{y}$ is not equal to $y$, the true label of that adversarial sample.

In both optimization problems above, $L(\cdot)$ represents the aforementioned cost function and $f(\cdot)$ denotes the DNN model trained with the traditional learning method discussed above . $\left \| \cdot  \right \|_{p}$ is $p$-norm -- sometimes also specified as $l_p$ distance -- indicating the dissimilarity between adversarial sample $\hat{x}$ and its corresponding legit data sample $x$. With different values of $p$ -- most popularly selected in adversarial learning research -- the optimization problems above can be computed in the following manners.

\begin{itemize}[leftmargin=*]

\item[(1)] With $p=2$, $p$-norm represents the measure of Euclidean distance. The constraint optimization problems above can be specified as unconstrained optimization problems
\begin{equation}
\label{eq:adv3}
\hat{x} = \, arg\, \underset{\hat{x}}{\mathrm{max}}\,L(f(\hat{x}; w); y) - c \cdot \left \| \hat{x}-x \right \|_{2}
\end{equation}
and
\begin{equation}
\label{eq:adv4}
\hat{x} = \, arg\, \underset{\hat{x}}{\mathrm{min}}\,L(f(\hat{x}; w); \hat{y}) + c \cdot \left \| \hat{x}-x \right \|_{2}.
\end{equation}
Here, both~\eqref{eq:adv3} and~\eqref{eq:adv4} can be solved by following either a first-order optimization method (e.g., stochastic gradient descent~\cite{Carlinioakland2017} and L-BFGS~\cite{42503}) or a second-order method (e.g., Newton-Raphson method).

\item[(2)] With $p=0$, $p$-norm indicates the number of elements in a legit data sample that an attacker needs to manipulate in order to turn it into an adversarial sample. Different from the computation method above, in the setting of $p=0$ where unconstrained optimization problems~\eqref{eq:adv3} and~\eqref{eq:adv4} are not differentiable, the computation for the optimal solution has to follow an approximation method introduced in~\cite{papernot2016limitations} or~\cite{Carlinioakland2017}. 

\item[(3)] With $p=\infty$, $p$-norm becomes a measure indicating the maximum change to individual features. As such, the optimal solution for~\eqref{eq:adv1} and~\eqref{eq:adv2} can be approximated by following the fast gradient sign method~\cite{Goodfellow14}, which computes perturbation $\partial L(f(x; w); y)/\partial x$ (or $\partial L(f(x; w); \hat{y})/\partial x$), multiplies it by distortion scale $\varepsilon$ and then adds the product to legitimate data sample $x$. Note that the way they compute that perturbation can be through back-propagation in conjunction with gradient descent.

\end{itemize}   

As is illustrated in the aforementioned optimization problems, in order to generate impactful adversarial samples, an attacker needs to know either standard DNN model $f(\cdot)$ or know of a way to approximate $f(\cdot)$. A recent study~\cite{42503} has revealed that an attacker could well approximate a standard DNN model using a traditional DNN training algorithm on an auxiliary training dataset. In this paper, we use ``{\em cross-model approach}'' to refer to those adversarial sample crafting methods that rely upon the approximation of a standard DNN model.

\commenta{

Here, $L(\cdot)$ and $f(\cdot)$ represent the aforementioned cost function and DNN model respectively. $\left \| \cdot  \right \|_{p}$ is $p$-norm, indicating the similarity between adversarial sample $\hat{x}$ and its corresponding legit data sample $x$. For simplicity, we denote $\left \| x-\hat{x}\right \|_{p}$ as the $l_{p}$ distance. To ensure distortion to a legit data sample is subtle, the aforementioned equations restrict distortion $l_{p}$ within a bound indicated by $\varepsilon$. 

Problem~\eqref{eq:adv1} and~\eqref{eq:adv2} are designed for generating adversarial samples with different attack purposes. More specifically,  by solving~\eqref{eq:adv1}, an adversary utilizes the generated adversarial sample $\hat{x}$ to cause $f(\cdot)$ to give arbitrarily wrong prediction with high confidence. While for~\eqref{eq:adv2}, the generated $\hat{x}$ is designed to lead $f(\cdot)$ to ``believe" $\hat{x}$ belongs to a specific incorrect class $y_i$. 

In order to solve either~\eqref{eq:adv1} or~\eqref{eq:adv2}, certain choice of $l_{p}$ must be made. Currently, the most prevalently used measures include $l_{0}$, $l_{2}$ and $l_{\infty}$: 

\begin{itemize}   
	\item[1)] $l_{0}$ measures the number of features of a sample that are modified by an attacker. \cite{papernot2016limitations, Carlinioakland2017} adopt $l_{0}$ for generating adversarial samples for images, and \cite{grosse2016adversarial} use $l_{0}$ to manipulate malware samples. Since $l_{0}$ distance is non-differentiable, \eqref{eq:adv1} and~\eqref{eq:adv2} using $l_{0}$ cannot be solved with gradient descent approach. Both \cite{papernot2016limitations} and~\cite{Carlinioakland2017} follow an iterative approach to generate adversarial samples heuristically. Specifically, at each iteration, \cite{papernot2016limitations} identify the most ``important'' pixel to manipulate, while~\cite{Carlinioakland2017}  pick the least important one and leave it fixed. The identification of their target pixels is realized by replacing $l_{0}$ with other differentiable measure (commonly $l_{2}$), then using either a saliency map~\cite{papernot2016limitations} or a gradient based approach~\cite{Carlinioakland2017} to evaluate the importance of pixels. After identifying the pixels to be altered, \cite{papernot2016limitations} and~\cite{Carlinioakland2017} manipulate these pixels by flipping their values from 0 to 1 or the opposite. Similar approach is also adopted in~\cite{grosse2016adversarial} for manipulating malware samples. 
	
	\item[2)]  $l_{2}$ measures the Euclidean distance between $x$ and $\hat{x}$, and has been used in~\cite{42503, Carlinioakland2017} to generate adversarial samples on image data sets. Since the original~\eqref{eq:adv1} and~\eqref{eq:adv2}  are difficult to solve, both~\cite{42503, Carlinioakland2017} instead solve $\hat{x} = \, arg\, \underset{\hat{x}}{\mathrm{max}}\,L(f(\hat{x}; w); y) - c \cdot \left \| \hat{x}-x \right \|_{2} $ (or $\hat{x} = \, arg\, \underset{\hat{x}}{\mathrm{min}}\,L(f(\hat{x}; w); y_i) + c \cdot \left \| \hat{x}-x \right \|_{2} $) for generating adversarial samples. These new formulations of the original problems can be solved by either a first-order optimization method (e.g., stochastic gradient descent~\cite{Carlinioakland2017} and L-BFGS~\cite{42503}) or a second-order method (e.g., Newton-Raphson method).
	
	\item[3)] $l_{\infty}$ measures the maximum change made to all features in $x$. \cite{Goodfellow14} first propose using $l_{\infty}$ to generate adversarial samples on image datasets. Similar to $l_{0}$, $l_{\infty}$ is also non-differentiable. \cite{Goodfellow14} develop a method named fast gradient sign, which computes perturbation $\partial L(f(x; w); y)/\partial x$ (or $\partial L(f(x; w); y_{i})/\partial x$) and then add it to legitimate sample $x$. The way they compute that perturbation can be through back-propagation in conjunction with gradient descent. It is clear to see that fast gradient sign solves neither~\eqref{eq:adv1} nor~\eqref{eq:adv2} to achieve optimal solution. As a result, it can be viewed as an efficient approximation method. \end{itemize}
}

\commenta{
More specifically, generating an adversarial sample can be viewed as solving the aforementioned optimization problem by computing perturbation $\partial^n L(f(\hat{x}; w); y)/\partial \hat{x}^n$ (or $\partial^n L(f(\hat{x}; w); y_{i})/\partial \hat{x}^n$) and then adding it to legitimate sample $x$. The way they compute that perturbation can be through back-propagation in conjunction with either a first-order optimization method (e.g., stochastic gradient descent and L-BFGS) or a second-order method (e.g., Newton-Raphson method), in which $n$ is equal to 1 and 2, respectively. 
}

\section{Existing Defenses and Problem Scope}
\label{sec:problem}

To counteract the adversarial learning problem described in the section above, recent research invent various training algorithms~\cite{Goodfellow14, ororbia_ii_unifying_2016, 42503, gu2014towards} to improve the robustness of a DNN model. They indicate, by using new training algorithms they design, one can improve a DNN's resistance to the adversarial samples crafted through the aforementioned cross-model approach. This is due to the fact that their training algorithms smooth a standard DNN's decision boundary, making adversarial samples -- impactful to standard DNN models -- no longer sufficiently effective. In this section, we summarize these defense mechanisms and discuss their limitations. Following our summary and discussion, we also define the problem scope of our research.

\subsection{Existing Defense Mechanisms}

Recently, research in hardening deep learning mainly focuses on two different tactics -- data augmentation and model complexity enhancement. Here, we summarize them in turn and disucss their limitations.

\noindent{\bf Data augmentation.} To resolve the issue of ``blind spots'' (a more informal name given to adversarial samples), many methods that could be considered as sophisticated forms of data augmentation\footnote{Data augmentation refers to artificially expanding the data-set. In the case of images, this can involve deformations and transformations, such as rotation and scaling, of original samples to create new variants.} have been proposed (e.g.~\cite{Goodfellow14,ororbia_ii_unifying_2016, gu2014towards}). In principle, these methods expand their training set by combining known samples with potential blind spots, the process of which has been called adversarial training~\cite{Goodfellow14}. Technically speaking, adversarial training can be formally described as adding a regularization term known as \emph{DataGrad} to a DNN's training loss function~\cite{ororbia_ii_unifying_2016}. The regularization penalizes the directions uncovered by adversarial perturbations (introduced in Section~\ref{sec:advlearn}). Therefore, adversarial training can work to improve the robustness of a standard DNN. 

\noindent{\bf Model complexity enhancement.} DNN models are already complex, with respect to both the nonlinear function that they try to approximate as well as their layered composition of many parameters. However, the underlying architecture is straightforward when it comes to facilitating the flow of information forwards and backwards, greatly alleviating the effort in generating adversarial samples. Therefore, several ideas~\cite{papernot2015distillation, gu2014towards} have been proposed to enhance the complexity of DNN models, aiming to improve the tolerance of complex DNN models with respect to adversarial samples generated from simple DNN models. For example, \cite{papernot2015distillation} developed a \emph{defensive distillation} mechanism, which trains a DNN from data samples that are ``distilled'' from another DNN. By using the knowledge transferred from the other DNN, the learned DNN classifiers become less sensitive to adversarial samples. Similarly, \cite{gu2014towards} proposed to stack an auto-encoder together with a standard DNN.  It shows that this auto-encoding enhancement increases a DNN's resistance to adversarial samples. 

\noindent{\bf Limitation.} While the aforementioned defenses have yielded promising results in terms of increasing model resistance, the scope of the model resistance they provide is relatively limited. Once an attacker obtains the knowledge of the new training algorithms -- instead of using a traditional DNN training algorithm to substitute the algorithm which the target DNN is trained with -- he can build his own model with the new training algorithm, and then use it as the cross model to facilitate the crafting of adversarial samples. As we will show in Section~\ref{sec:eval}, the adversarial samples crafted through such new cross models sustain their offensiveness to the corresponding DNN models. This indicates that the effectiveness of existing defense mechanisms is highly dependent upon the obscurity of training algorithms. 

\subsection{Problem Scope}

With the existing defenses and their limitation in mind, here we define the problem scope of our research. 

Similar to most previous research -- if not all -- in hardening deep learning, we assume that an attacker crafts adversarial samples by solving the aforementioned optimization problems with derivative calculation (e.g., fast sign gradient descent or Newton-Raphson method). We believe this assumption is realistic for the following reason. 

Derivative calculation is the most general approach for solving an optimization problem. In the future, while one might be able to derive new forms of approaches in solving the aforementioned optimization problem, he or she has to ensure the new approaches are computationally efficient. Without the aid from derivative calculation, this can be relatively difficult. Even if one could computationally efficiently resolve the aforementioned optimization problems -- for example perhaps through relaxation -- without derivative calculation, he still need to prove the adversarial samples derived from such an approach are impactful. Given that relaxation reshapes an optimization problem, the ``optimal'' solution may not even close to any local optima of that original optimization problem. 

\begin{figure*}[t]
    \begin{center}
        \includegraphics[scale=0.45]{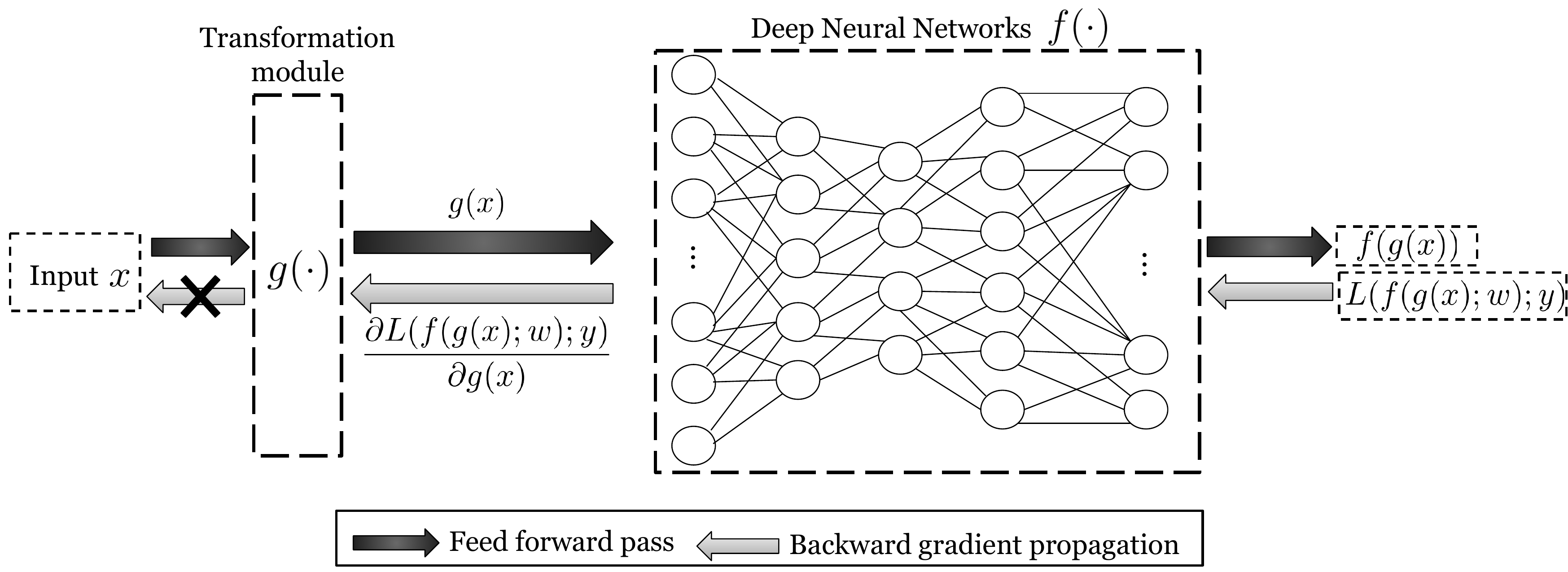}\\
        \caption{A deep neural network with a data transformation module, projecting $X$, an input data sample, to new representation $g(X)$ prior to passing it through the DNN, $f(\cdot)$.}
        \label{fig:frame}
    \end{center}
\end{figure*}

Different from prior research, we also assume that an adversary has not only the access to a DNN's structure as well as the dataset(s) used to build the DNN but more importantly the algorithm used to train the network. In other words, we assume a target DNN model is no longer obscure to an adversary and rather he has the full knowledge about a DNN model that he attempts to exploit. We believe this assumption is more practical because there is little hope of keeping an adversary-resistant training algorithm completely secret from dedicated attackers. In the long run, any system the security of which relies upon the obscurity of its design can be doomed~\cite{Kerckhoffs1883}.


\commenta{

As is discussed in Section~\ref{}, adversarial samples exist in a relative large subspace, and there is no way to build a DNN model that guarantees in covering all possible blind-spots. Similar to the past research in adversarial learning, the ultimate goal of this work therefore is to increase the difficulty for attackers in finding adversarial samples. More specifically, this work focuses on developing an adversary-resistant learning algorithm, comparable to the existing approaches in terms of defending against adversarial samples crafted through the cross-model scheme. 

when our training algorithm is disclosed, make an attacker not capable of crafting adversarial samples through the optimization methods discussed in Section~\ref{}.  

Making a DNN model at least remain the same level of resistance to adversarial samples, when an attacker attempts to use the same strategies to compromise the DNN model. In other words, comparable results ...

}

\section{Our Approach}
\label{sec:framework}

To address the problem above, we propose a new approach to harden DNN models. Technically speaking, it follows the tactic of model complexity enhancement, which improves model resistance by increasing model complexity. Different from the existing techniques mentioned above, our approach however goes beyond the scope of robustness they provide. It ensures that an attacker cannot perform the aforementioned attack to generate adversarial samples impactful to our learning model even if we reveal our training algorithm. In other words, our approach escalate a DNN's resistance to adversarial samples without the requirement of obscuring training algorithms.

As is discussed in the section above, the adversarial learning problem can be viewed as an optimization problem. To resolve that optimization problem, one needs to conduct analytical computation of gradients with respect to an input data sample and perform backward propagation accordingly. Therefore, we escalate a DNN's robustness not only by increasing the model complexity but, more importantly, restricting back-propagation. 

More specifically, we integrate to a DNN a data transformation module, $g(\cdot)$ graphically indicated in Figure~\ref{fig:frame}. As is illustrated in the figure, the data transformation module projects $X$, an input data sample to $g(X)$, a new representation, before passing it through a consecutive DNN. This transformation increases the complexity of a DNN model and augments its resistance to adversarial samples crafted through the aforementioned cross-model scheme. In addition, it blocks the backward flows of gradients. With this block, even if the underlying training algorithm is disclosed, an adversary cannot craft adversarial samples. In the following, we specify the design principle of our data transformation module, followed by its design detail and some necessary discussions.

\subsection{Design Principle}
\label{sec:principle}

To block the backward flow of gradients, the design of data transformation must satisfy three requirements. Most notably, the data transformation must be non-differentiable. As is discussed above, crafting adversarial samples requires the calculation of gradients as well as the back-propagation of those gradients. By making data transformation module $g(\cdot)$ non-differentiable, therefore, we can make gradient calculation intractable and thus obstruct the backward flow of gradients. More formally, we can choose non-differentiable function $g(x)$, making the derivative difficult to be calculated, i.e.,
\begin{equation}
\frac{\partial^n L(f(g(\hat{x}); w); y)}{\partial \hat{x}^n} = \frac{\partial^n L(f(g(\hat{x}); w); y)}{\partial g(\hat{x})^n} \cdot \frac{\partial^n g(\hat{x})}{\partial \hat{x}^n}.
\end{equation}
Here, $f(\cdot)$ represents the DNN model in tandem with the data transformation module, and $L(\cdot)$ denotes the cost function described in Section~\ref{sec:advlearn}. The derivative can be computed using either a first-order optimization method (e.g., gradient descent) or a second-order method (e.g., Newton-Raphson method), in which $n$ is equal to 1 and 2, respectively.

While the non-differentiability feature restricts the crafting of adversarial samples, an adversary might still be able to generate adversarial samples. Since end-to-end gradient flow is blocked at the input layer of the successive DNN, back-propagation can only carry error gradients to the output of the transformation module. Given $g(\cdot)$, an adversary could construct an adversarial sample by inverting transformation module $g(\cdot)$ and passing the manipulated transformation output through the inversion of the transformation. More formally, the adversary can construct an adversarial sample by computing 
\begin{equation}
g^{-1}(g(\hat{x}) + \frac{\partial^n L(f(g(\hat{x}); w); y)}{\partial g(\hat{x})^n}).
\end{equation} 
In addition to making data transformation non-differentiable, therefore, we must further ensure that the inversion of the data transformation is computationally intractable. In other words, the data transformation $g(\cdot)$ needs to have the property of non-invertibility.

Satisfying the two requirements above ensures that our proposed approach can prohibit an attacker from crafting adversarial samples directly from the target DNN model, and one does not need to concern about the disclosure of training algorithms. However, the data transformation proposed may significantly jeopardize the accuracy of a DNN model if not designed carefully. Take the following extreme case for example. 

Hash functions like MD5 and SHA1 are one-way functions which have the properties of non-differentiability as well as non-invertibility. By simply using it as the transformation module, we can easily prohibit an attacker from crafting adversarial samples even if he knows of which hash function we choose and how we integrate it with DNNs. However, a hash function significantly changes the distribution of input data samples. Armed with it, a DNN model suffers from significant loss in classification performance. Last but not least, our design therefore must ensure data transformation can preserve the distribution of data representation. This can potentially make a DNN robust without sacrificing classification performance.

\subsection{Design Detail}

Following the design principle above, we choose Locally Linear Embedding (LLE)~\cite{roweis2000nonlinear}, a non-parametric dimensionality reduction mechanism, serving as the data transformation module. As we will discuss in the following, this representative non-parametric method is non-differentiable. More importantly, it can be theoretically proven that inverting LLE is an NP-hard problem. Last but not least, LLE seeks low-dimensional, neighborhood-preserving map of high-dimensional input samples, and thus is a method that best suited to preserving as much information in the input as possible. In the following, we first describe LLE and then expound upon the fact that, as a non-parametric dimensionality reduction method, LLE is non-differentiable. Furthermore, we theoretically prove LLE is computationally non-invertible. 

\begin{figure*}[t]
    \centering
    \begin{subfigure}[t]{0.43\textwidth}
        \includegraphics[width=1.0\textwidth]{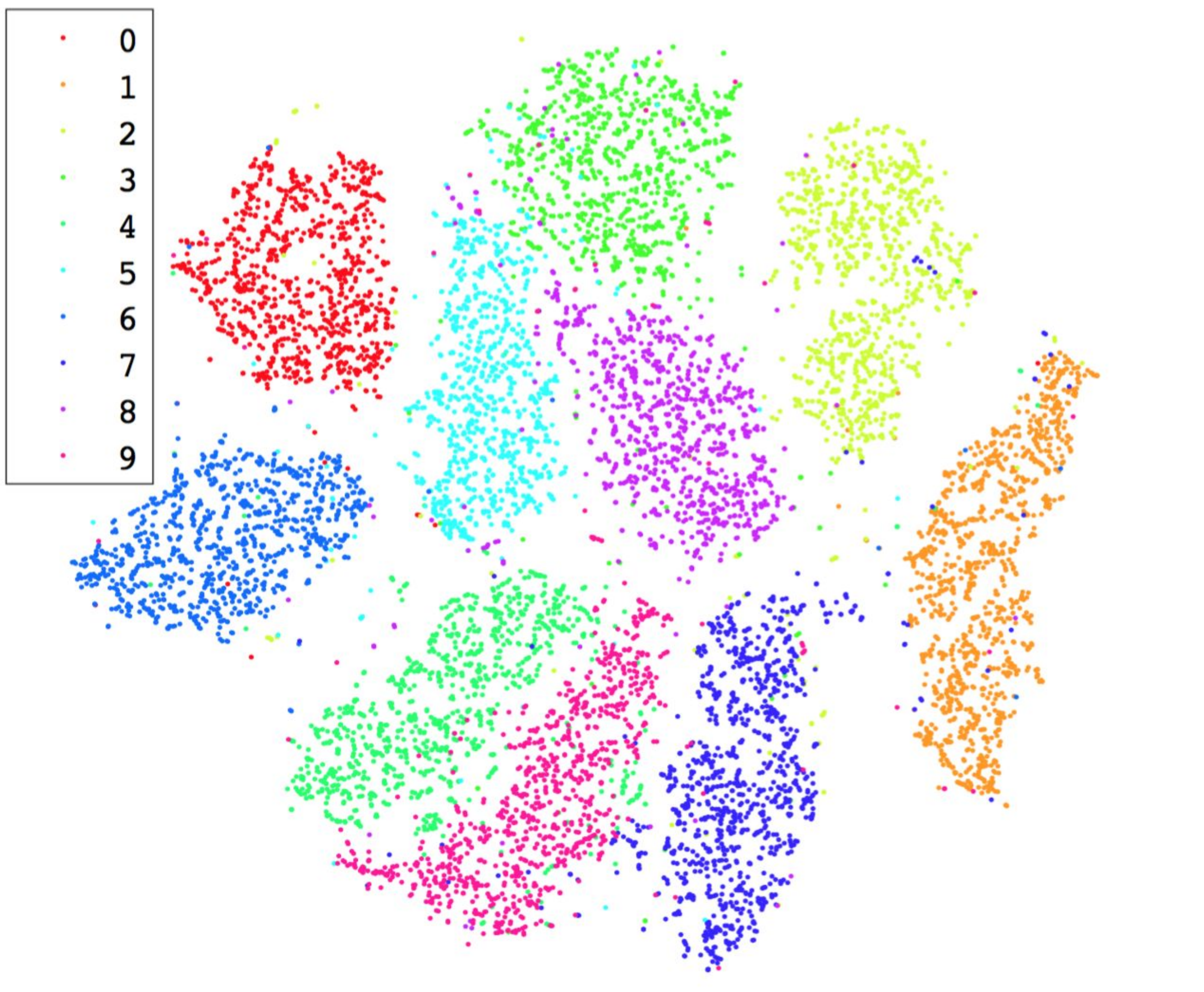}
        \caption{Before LLE processing.}
        \label{subfig:MNIST_org1}
    \end{subfigure}
    ~
    \begin{subfigure}[t]{0.43\textwidth}
        \includegraphics[width=1.0\textwidth]{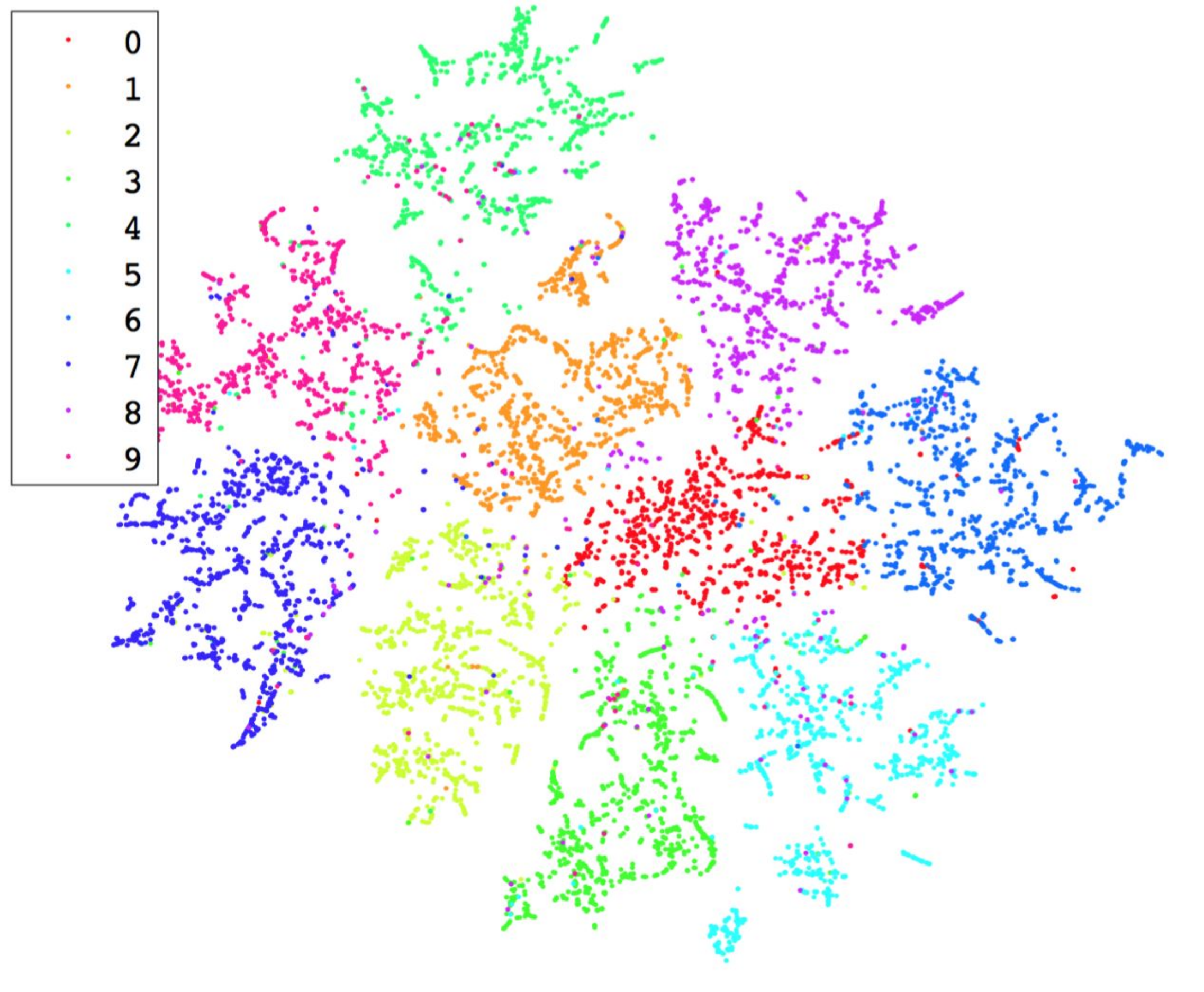}
        \caption{After LLE processing.}
        \label{subfig:MNIST_LLE1}
    \end{subfigure}
    \caption{Visualization of the MNIST data set~\cite{lecun1998mnist} in a 2-dimensionality space before and after proceeded by LLE.}
    \label{fig:visual_mnist}
\end{figure*}

\subsubsection{Locally Linear Embedding}

LLE is a non-parametric method designed to reduce input data dimensionality and at the same time preserve local properties of high-dimensional input in a lower-dimensional space. To some extent, this can ensure the distribution of high-dimensional data samples is as close as they are in a lower-dimensional space. Technically speaking, this is achieved by representing each high-dimensional data sample via a linear combination of its nearest neighbors. More formally, this can be expressed as $x_{i} = \sum_{j=1}^{N} w_{ij} \cdot x_{j}$. Here, $x_i$ and $x_{j}$ ($x_{i}, x_{j} \in \mathbb{R}^{1 \times m}$) denote the $i^{th}$ data sample and its $j^{th}$ neighbor ($j = 1, 2..., N$), respectively. $w_{ij}$ represents the weight, indicating the contribution of $x_{j}$ to data sample $x_{i}$. As is described in~\cite{roweis2000nonlinear}, those weights (a.k.a. reconstruction weights) can be represented as weight matrix $W$ and computed by solving the following optimization problem:
\begin{equation}
    \begin{aligned}
    \label{eq:lle_1}
        &\underset{W}{\mathrm{min}} \hspace{0.2cm} \sum_{i} \big \| x_{i}-\sum_{j}w_{ij} \cdot x_{j} \|_2^{2}\\
        &\mathrm{s.t.} \hspace{0.3cm} \sum_{j}w_{ij}=1.
    \end{aligned}
\end{equation}
In weight matrix $W$, LLE deems $w_{ij}=0$ if $x_{j}$ is not considered as a neighbor of $x_{i}$, and the total number of neighbors assigned to $x_{i}$ is a carefully selected hyper-parameter. The neighboring relation between $x_{i}$ and $x_{j}$ depends on the value of the $l_2$ distance between $x_{i}$ and $x_{j}$.

Since the reconstruction weights encode the local properties of the high-dimensional data, they can be used to preserve the data distribution at the time of performing dimensionality reduction. More specifically, LLE imposes the corresponding reconstruction weights to each lower-dimensional data sample via a similar linear combination, and then attempts to find $Y = \{y_{1}, y_{2}, \dots, y_{N} \}$, the lower-dimensional representation of $X = \{x_{1}, x_{2}, \dots, x_{N}\}$ by solving the following optimization problem: 
\begin{equation}
    \begin{aligned}
    \label{eq:lle_2}
        &\underset{Y}{\mathrm{min}} \hspace{0.2cm}\sum_{i}\big \| y_{i}-\sum_{j}w_{ij} \cdot y_{j}  \|_2^{2} \\
        &\mathrm{s.t.} \hspace{0.3cm}\sum_{i}y_{i}=0,\\
              &\hspace{0.8cm} \frac{1}{N}\sum_{i}y_{i}^\mathrm{T}y_{i} = I.
    \end{aligned}
\end{equation}
where $y_{i}, y_{j} \in R^{1 \times m_c}$, indicating $y_i$, $y_j$ consist of $m_{c}$ of elements.

In order to solve the optimization problem above, the Rayleitz-Ritz theorem~\cite{horn1990matrix} is typically used. It computes the eigenvectors corresponding to the smallest nonzero eigenvalues of the inner matrix product $(I-W)^{\mathrm{T}} \cdot (I-W)$. For a detailed explication, we refer the readers to~\cite{horn1990matrix}. Here, $I \in \mathbb{R}^{N \times N}$ is an identity matrix, and $W \in \mathbb{R}^{N \times N}$ is the aforementioned reconstruction weight matrix.  

LLE is specifically designed to retain the similarity between pairs of high dimensional samples when they are mapped to lower dimensions~\cite{roweis2000nonlinear}. To illustrate this property, we provide a visualization of the data before and after being processed by LLE in Figure~\ref{fig:visual_mnist}. The visualization result demonstrates that using LLE as a data transformation module satisfies the last design principle discussed in Section~\ref{sec:principle} (i.e., preserving the distribution of the original data). More importantly, this property also helps bound the lower dimensional mapping of adversarial samples to a vicinity which is filled by mappings of original test samples that are highly similar to these adversarial samples. As a result, there is a significantly lower chance that an adversarial sample acts as an outliers in the lower dimensional space. In other words, LLE makes a DNN more resistant to adversarial samples. In Section~\ref{sec:eval}, we empirically validate this important property.

\subsubsection{Non-differentiability of LLE}

Existing dimensionality reduction methods can be categorized as either parametric or non-parametric~\cite{Maaten08dimensionality}. Parametric methods utilize a fixed amount of parameters to specify a direct mapping from high-dimensional data samples to their low-dimensional projections (or vice versa). This direct mapping is characterized by  parameters, which are typically optimized to provide the best mapping performance. This is similar to the functionality provided by a standard DNN, which maps high-dimensional data samples to the final decision space through differentiable function $f(\cdot)$. As such, the derivative of parametric methods typically can be computed in an analytically efficient manner. In other words, parametric methods are generally differentiable, and this nature becomes a disadvantage for blocking the backward gradient flow. 

On the contrary, non-parametric methods do not suffer from the issue above. For any non-parametric method, $g(\cdot)$, there is no way to express it in a closed form. Therefore, the derivative of $g(\cdot)$ can be computed only through a numeric but not an analytical approach. More formally, this means the calculation of $\partial g(x)/\partial x$ needs to be completed through the calculation of limit $\lim_{h \rightarrow 0} \frac{g(x+h)-g(x)}{h}$. Given that a deep neural network takes as input each individual sample, which is discrete in the sample space, it is difficult to define the continuity of $g(\cdot)$ with traditional topology and thus the differentiability of $g(\cdot)$ cannot be guaranteed. This indicates, as a member of non-parametric methods, LLE perfectly satisfies the first design principle discussed in Section~\ref{sec:principle} (i.e., not capable of performing derivative calculation).

\subsubsection{Non-invertibility of LLE}

We validate the non-invertibility of LLE by theoretically proving that reconstructing original high-dimensional data from low-dimensional representations transformed by LLE is computationally intractable. More formally, we prove that, given a set of low-dimensional data $Y=\{y_{1}, y_{2}, \dots, y_{N}\}$ ($Y \in \mathbb{R}^{N\times m_c}$) produced by LLE, reconstructing their original high-dimensional representations $X=\{x_{1}, x_{2}, \dots, x_{N}\}$ ($X \in \mathbb{R}^{N\times m}$) from $Y$ is at least an NP-hard problem.

Recall that LLE computes weight matrix $W$ and utilizes it to project high-dimensional data samples to a lower-dimensional space. As a result, to restore high-dimensional data from its lower-dimensional representations, one has to recover that matrix by following the calculation similar to that shown in~\eqref{eq:lle_1}, except that $x_{i}$ and $x_{j}$ are replaced by $y_{i}$ and $y_{j}$. 

Once weight matrix $W$ is restored, the recovery of original high-dimensional data can be viewed as solving the following optimization problem:
\begin{equation}
\begin{aligned}
\label{eq:relle_1}
    \underset{X}{\mathrm{min}} \hspace{0.2cm}\sum_{i}\big \| x_{i}-\sum_{j}{w}_{ij} \cdot x_{j} \big \|_2^{2}.
\end{aligned}
\end{equation}
It is not difficult to realize that Equation~\eqref{eq:relle_1} can be defined in the following quadratic form:
\begin{equation}
  \begin{aligned}
  \label{eq:relle_2}
    \underset{X}{\mathrm{min}} \hspace{0.2cm}\sum_{i,j} m_{ij} \cdot (x_{i} \cdot x_{j}),
  \end{aligned}
\end{equation}
where $m_{ij} = \delta_{ij} - {w}_{ij} -  {w}_{ji} + \sum_{k} {w}_{ki} {w}_{kj}$. Note that $\delta_{ij}=1$ if $i=j$ and $0$ otherwise. If expressing $m_{ij}$ as the following matrix:
\begin{equation}
M = 
 \begin{pmatrix}
  m_{11} & m_{12} & \cdots & m_{1N} \\
  m_{21} & m_{22} & \cdots & m_{2N} \\
  \vdots  & \vdots  & \ddots & \vdots  \\
  m_{N1} & m_{N2} & \cdots & m_{NN}
 \end{pmatrix},
\end{equation}
we can easily realize that $M$ is a symmetric matrix, which has the property of $M^\mathrm{T} = M$. 

Now, with the analysis above, the validation of non-invertibility amounts to proving that solving (\ref{eq:relle_2}) is at least an NP-hard problem. In this work, we conduct this proof by introducing several constraints to this equation. Our basic idea is to use these constraints to relax the optimization problem in (\ref{eq:relle_2}) to a nearby problem which can be easily proved as an NP-hard problem. More specifically, we introduce the following constraints: 
\begin{equation}
\sum_{i}x_{i}=\vec{0},
\label{eq:con_1}
\end{equation}
and
\begin{equation}
-1 \leq x_{ij} \leq 1, \forall i \in N, j \in m,
\label{eq:con_2}
\end{equation}
where $\vec{0}$ denotes a zero vector, and $x_{ij}$ represents the $j^{th}$ element in vector $x_i$. 

With these constraints introduced to Equation~\ref{eq:relle_2}, we can relax the optimization problem to a quadratic problem with a non-positive semi-definite constraint, which itself is a class of NP-hard problems~\cite{d2003relaxations}. In the following, we provide more details on why the involvement of the aforementioned constraints transforms the optimization problem in (\ref{eq:relle_2}) to this class of NP-hard problems.  
        
Let $ A\in \mathbb{R}^{N_A\times 1}$ denote a column vector which is the concatenation of $x_{i}$ for $i = 1, \dots,N$ and $N_A = N \times m$. Then, we have $A=(x_{1},x_{2},\cdots ,x_{N})^{\mathrm{T}}$. Let $q\in \mathbb{R}^{1\times N_A}$ denote a row vector in which every element is equal to $1$. We further define matrices $P,Q\in \mathbb{R}^{N_A\times N_A}$ as follows:
\begin{equation}
\begin{aligned}
\label{eq:concatenation}
    &P=\begin{pmatrix}
    P_{11} & P_{12}&\cdots \;P_{1N} \\ 
    P_{21} & P_{22} & \cdots\;P_{2N}\\ 
    \vdots & \vdots & \quad \vdots\\
    P_{N1}&P_{N2} &\cdots \;P_{NN}
    \end{pmatrix}, \quad Q=-I,
\end{aligned}
\end{equation}
where $P_{ij}=m_{ij} \cdot J$. $J \in \mathbb{R}^{m\times m}$ is a matrix of ones where every element is equal to $1$. 

Given the constraint in (\ref{eq:con_1}) that we introduce, it is not difficult to discover $\sum_{i}x_{i}^{T}=\vec{0}$. Since the multiplication of a vector and its transpose derives a non-negative value, we have $\Sigma_{i}x_ix_i^T \geq 0$ and the constraint in (\ref{eq:con_1}) can be expressed as inequation $-\Sigma_{i}x_ix_i^T +\sum_{i}x_{i}^T+\alpha \leq 0$, indicating there exists a positive number, $\alpha$ that always holds the inequity. By rewriting the inequation using the notations newly defined above, we can therefore transform the constraint in (\ref{eq:con_1}) into the form of $A^{\mathrm{T}}QA+qA+\alpha\leq 0$.

Given the constraint in \eqref{eq:con_2}, we can easily derive inequation $\Sigma_{i}x_ix_i^T-N_A \leq 0$, which can be further expressed as $\Sigma_{i}x_ix_i^T-N_A+\gamma \leq 0$ indicating there alway exists a constant, $\gamma$ that holds the inequity. By rewriting both the constraint itself and this inequation using newly defined notations, we can derive constraints  $\left \| A \right \|_{\infty} \leq 1$ as well as $A^{\mathrm{T}}IA-N_A+\gamma \leq 0$. As such, we can transform Equation~\eqref{eq:relle_2} and the aforementioned constraints in~\eqref{eq:con_1} and~\eqref{eq:con_2} into following form:
\begin{equation}
\begin{aligned}
\label{eq:QCQP}
    &\mathrm{min}  \hspace{0.2cm} A^{\mathrm{T}}PA\\
    &\mathrm{s.t.} \hspace{0.3cm} A^{\mathrm{T}}QA+qA+\alpha\leq 0.\\
    &\hspace{0.8cm} A^{\mathrm{T}}IA-N_A+\gamma \leq 0.\\
    &\hspace{0.8cm} \left \| A \right \|_{\infty} \leq 1.
\end{aligned}
\end{equation}   
Here, $Q$ is negative semi-definite, and thus Equation~\eqref{eq:QCQP} is a quadratic problem with a non-positive semi-definite constraint. According to~\cite{Vavasis1991Nonlinear, d2003relaxations}, Equation~\eqref{eq:QCQP} belongs to a class of NP-hard problems, which implies the non-invertability of LLE.

\subsection{Discussion}

Here, we discuss some related issues and possible attacks against our proposed technique.

\noindent
{\bf Approximation of LLE.} 
While the aforementioned discussion and theoretical proof have already indicated the effectiveness of our proposed approach, intuition suggests that an adversary might still come up with an attack. Specifically, he might approximate LLE using a parametric mapping and then substitute LLE accordingly. Since parametric mappings do not have the property of non-differentiability, the adversary can take advantage of the substitute, pass gradients through and eventually craft adversarial samples. However, as we will show in Section~\ref{sec:eval}, even using the state-of-the-art approximation scheme, an adversary cannot craft impactful adversarial samples.

\noindent
{\bf Other dimensionality reduction methods.}
As is described above, we choose LLE, a representative non-parametric dimensionality reduction method, to serve as the data transformation module. This is due to the fact that it provides many properties needed for hardening a DNN, such as non-differentiability, non-invertibility and the capability of preserving data distribution. 

Going beyond LLE, there are other non-parametric dimensionality reduction methods that offer the same properties, e.g., t-Distributed Stochastic Neighbor Embedding (t-SNE)~\cite{maaten2008visualizing} and Sammon Mapping~\cite{sammon1969nonlinear}. However, they cannot be utilized in our problem domain for the following reason. 

Deep neural networks exhibit superior performance when dealing with data in a relatively high dimensionality. Other non-parametric methods are typically designed more for tasks like visualization~\cite{maaten2008visualizing} where it is required that the dimensionality of the mappings is two or three. Using them as our data transformation module, they cannot provide high-dimensional data input for the DNN in tandem with the transformation, and may significantly jeopardize classification performance.

\commenta{
Recall our proof that LLE, as a non-parametric dimensionality reduction method, can block the end-to-end gradient flow. It suffices to show that an architecture, built from a non-parametric data transformation module, such as LLE, followed by a standard DNN is theoretically guaranteed to be resistant to white box adversarial samples.
}

\commenta{
Considering that -- without relaxation -- the optimal solution for Equation~\ref{eq:relle_2} cannot be determined (see Appendix), we conduct our proof by introducing the following constraints to this equation. By examining Equation~\ref{eq:relle_1}, the non-quadratic form of Equation~\ref{eq:relle_2}, we can easily discover the non-quadratic form equation shares a similar cost function with Equation~\ref{eq:lle_2}. Therefore, intuition suggests we can introduce both constraints associated with that equation.

It is natural to consider adopting the two constraints in~\eqref{eq:lle_2}. In particular, the first constraint $\sum_{i}y_{i}=0$ can be similarly applied to~\eqref{eq:relle_2} without affecting the cost. In our case, without loss of generality, we assume that each element in $z_{i}$ resides in the range from -1 to 1. As for the second constraint,  $\frac{1}{N}\sum_{i}y_{i}^\mathrm{T}y_{i} = I$, it is noted that $Y$ is column-wise orthogonal. This implies that $Y$ represents original data perfectly without containing redundancy. As a result, this second constraint cannot be applied to~\eqref{eq:relle_2}, as there is no way of mapping a low-dimensional representation to a high-dimensional space while maintaining the property of column-wise orthogonality. Indeed, upon closer examination of $X$, one may observe that $X$ can be constructed from $Y$ by taking the columns of $Y$ as the basis for the column space of $X$. This further implies that there exists a correlation between any pair of columns of $X$. Given above analysis, we set a constant $\gamma$ as data redundancy of high-dimensional data and propose two constraints for solving~\eqref{eq:relle_1}
}

\commenta{
Until now there have been no clear constraints imposed on ~\eqref{eq:relle_2}. Therefore, in what follows, we first consider solving~\eqref{eq:relle_2} by assuming that there are no constraints. 

Problem~\eqref{eq:relle_2} is equal to $R(X) = X^{\mathrm{T}}MX$ if we transform~\eqref{eq:relle_2} by matrix multiplication. It then follows that the derivative of $R(X)$ with respect to $X$ is:
\begin{equation}
  \begin{aligned}
  \label{eq:relle_3}
    \frac{\mathrm{d} R(X)}{\mathrm{d} X} &= X^\mathrm{T} M \frac{\mathrm{d} X}{\mathrm{d} X}+ X^{\mathrm{T}}  M^{\mathrm{T}}  \frac{\mathrm{d}X}{\mathrm{d} X}\\
    &= 2X^\mathrm{T} M.
  \end{aligned}
\end{equation}
Equation \eqref{eq:relle_3} is obtained under the condition $M^\mathrm{T} = M$. The optimal $X$ for minimizing $R(X)$ can be calculated when $\mathrm{d} R(X)/\mathrm{d} X = 0$. This further leads to:

\begin{equation}
  \begin{aligned}
  \label{eq:solveX}
    2X^\mathrm{T}M = 0.
  \end{aligned}
\end{equation}
 
Note that the solution of the linear equation \eqref{eq:solveX} depends on the rank($M$). If rank($M$)$=$$N$, \eqref{eq:solveX} has a unique solution, $X=0$. Otherwise, if rank($M$)$<$$N$, the optimal solution for \eqref{eq:solveX} cannot be determined without any other constraint. This is obviously not a desirable method when considering the practical case. Therefore, assuming that there are no constraints for~\eqref{eq:relle_2} is invalid.
}

\begin{table*}[t]
\centering
\begin{tabular}{ccccccccccccc}
\hline
\multirow{3}{*}{\begin{tabular}[c]{@{}c@{}}Learning \\ Technology\end{tabular}} & \multicolumn{6}{c|}{Black Box}                                                                                                      & \multicolumn{6}{c}{White Box}                                                                                   \\ \cline{2-13} 
                                                                                & \multicolumn{3}{c|}{MNIST}                             & \multicolumn{1}{c|}{MALWARE} & \multicolumn{2}{c|}{IMDB}                   & \multicolumn{3}{c|}{MNIST}                             & \multicolumn{1}{c|}{MALWARE} & \multicolumn{2}{c}{IMDB} \\ \cline{2-13} 
                                                                                & $l_{\infty}$ & $l_{2}$ & \multicolumn{1}{c|}{$l_{0}$} & \multicolumn{1}{c|}{$l_{0}$} & $l_{\infty}$ & \multicolumn{1}{c|}{$l_{2}$} & $l_{\infty}$ & $l_{2}$ & \multicolumn{1}{c|}{$l_{0}$} & \multicolumn{1}{c|}{$l_{0}$} & $l_{\infty}$  & $l_{2}$  \\ \hline
Standard DNN                                                                    & 6.86\%       & 6.40\%  & \multicolumn{1}{c|}{7.50\%}  & \multicolumn{1}{c|}{26.19\%} & 28.10\%      & \multicolumn{1}{c|}{29.56\%} & 6.86\%       & 6.40\%  & \multicolumn{1}{c|}{7.50\%}  & \multicolumn{1}{c|}{26.19\%} & 28.10\%       & 29.56\%  \\ \hline
Distillation                                                                    & 87.06\%      & 96.22\% & \multicolumn{1}{c|}{47.36\%} & \multicolumn{1}{c|}{79.93\%} & 82.65\%      & \multicolumn{1}{c|}{87.31\%} & 34.43\%      & 12.60\% & \multicolumn{1}{c|}{8.43\%}  & \multicolumn{1}{c|}{40.47\%} & 48.98\%       & 49.12\%  \\ \hline
Adv. Training                                                                   & 89.09\%      & 96.23\% & \multicolumn{1}{c|}{84.33\%} & \multicolumn{1}{c|}{96.70\%} & 87.43\%      & \multicolumn{1}{c|}{87.66\%} & 33.94\%      & 14.44\% & \multicolumn{1}{c|}{8.89\%}  & \multicolumn{1}{c|}{43.09\%} & 50.78\%       & 51.0\%   \\ \hline
LLE-DNN                                                                         & 95.25\%      & 96.59\% & \multicolumn{1}{c|}{86.12\%} & \multicolumn{1}{c|}{95.02\%} & 87.58\%      & \multicolumn{1}{c|}{87.69\%} & 97.02\%      & 97.45\% & \multicolumn{1}{c|}{87.49\%} & \multicolumn{1}{c|}{94.66\%} & 87.47\%       & 87.53\%  \\ \hline
\end{tabular}
\caption{The comparison of model resistance to adversarial samples crafted in different manners. The values in the table represent the classification accuracy that DNN models exhibit when classifying adversarial samples.}
\label{tab:resistance}
\end{table*}

\begin{table}[]
\centering
\begin{tabular}{ll}
\tabucline[2pt]{-}
\multicolumn{2}{c}{\multirow{2}{*}{ Feature Examples}}                                                                                                                 \\
\multicolumn{2}{c}{}                                                                                                                                          \\ \tabucline[2pt]{-}
WINDOWS\_FILE:Execute:{[}system{]}$\setminus$slc.dll \\ \hline 
WINDOWS\_FILE:Execute:{[}system{]}$\setminus$cryptsp.dll \\ \hline
\multicolumn{2}{l}{\begin{tabular}[c]{@{}l@{}}
WINDOWS\_FILE:Execute:{[}system{]}$\setminus$slc.dll, \\ 
WINDOWS\_FILE:Execute:{[}system{]}$\setminus$cryptsp.dll \\
WINDOWS\_FILE:Execute:{[}system{]}$\setminus$wersvc.dll, \\ 
\end{tabular}}     \\ \hline
WINDOWS\_FILE:Execute:{[}system{]}$\setminus$faultrep.dll \\ \hline
\multicolumn{2}{l}{\begin{tabular}[c]{@{}l@{}}WINDOWS\_FILE:Execute:{[}system{]}$\setminus$imm32.dll, \\ WINDOWS\_FILE:Execute:{[}system{]}$\setminus$wer.dll\end{tabular}}       \\ \hline
\multicolumn{2}{l}{\begin{tabular}[c]{@{}l@{}}WINDOWS\_FILE:Execute:{[}system{]}$\setminus$ntmarta.dll, \\ WINDOWS\_FILE:Execute:{[}system{]}$\setminus$apphelp.dll\end{tabular}} \\ \hline
WINDOWS\_FILE:Execute:{[}fonts{]}$\setminus$times.ttf \\ \tabucline[2pt]{-}
\end{tabular}
\caption{An illustration of features in malware dataset. Individual features are represented in rows. They are either a filesystem access or a sequence of filesystem events.}
\label{tab:malware}
\end{table}

\begin{table*}[t]
\centering
\begin{tabular}{ccccccccc}
\hline

                                                                                  & \multicolumn{1}{l}{}                           & \multicolumn{7}{c}{Hyper Parameters}                                                                                                                                                                                                          \\ \cline{3-9} 
\multirow{-2}{*}{\begin{tabular}[c]{@{}c@{}}Training \\ Algorithms\end{tabular}}  & \multicolumn{1}{l}{\multirow{-2}{*}{Datasets}} & DNN Structure                                 & \multicolumn{1}{l}{Activation}  & Optimizer                       & Learning Rate                 & Dropout                       & Batch                       & Epoch                       \\ \hline
                                                                                  & MNIST                                          & 784-500-300-100                               & Sigmoid                         & Adam                            & 0.001                         & -                             & 100                         & 70                          \\
                                                                                  & \cellcolor[HTML]{EFEFEF}Malware                & \cellcolor[HTML]{EFEFEF}3738-3000-1000-100-2 & \cellcolor[HTML]{EFEFEF}Relu    & \cellcolor[HTML]{EFEFEF}Adam & \cellcolor[HTML]{EFEFEF}0.001 & \cellcolor[HTML]{EFEFEF}0.25 & \cellcolor[HTML]{EFEFEF}500 & \cellcolor[HTML]{EFEFEF}20  \\
\multirow{-3}{*}{\begin{tabular}[c]{@{}c@{}}Standard \\ DNN\end{tabular}}         & \cellcolor[HTML]{C0C0C0}IMDB                   & \cellcolor[HTML]{C0C0C0}600-200-200-100-2     & \cellcolor[HTML]{C0C0C0}Tanh    & \cellcolor[HTML]{C0C0C0}Adam    & \cellcolor[HTML]{C0C0C0}0.001 & \cellcolor[HTML]{C0C0C0}0.5   & \cellcolor[HTML]{C0C0C0}100 & \cellcolor[HTML]{C0C0C0}40  \\ \hline
                                                                                  & MNIST                                          & 784-100-100-100-10                            & Tanh                            & SGD                             & 0.1                           & 0.25                          & 100                         & 60                          \\
                                                                                  & \cellcolor[HTML]{EFEFEF}Malware                & \cellcolor[HTML]{EFEFEF}3738-3000-1000-100-2 & \cellcolor[HTML]{EFEFEF}Relu    & \cellcolor[HTML]{EFEFEF}Adam & \cellcolor[HTML]{EFEFEF}0.001 & \cellcolor[HTML]{EFEFEF}0.25 & \cellcolor[HTML]{EFEFEF}500 & \cellcolor[HTML]{EFEFEF}20  \\
\multirow{-3}{*}{\begin{tabular}[c]{@{}c@{}}Adversarial \\ Training\end{tabular}} & \cellcolor[HTML]{C0C0C0}IMDB                   & \cellcolor[HTML]{C0C0C0}600-300-100-50-2      & \cellcolor[HTML]{C0C0C0}Sigmoid & \cellcolor[HTML]{C0C0C0}Adam    & \cellcolor[HTML]{C0C0C0}0.001 & \cellcolor[HTML]{C0C0C0}0.2   & \cellcolor[HTML]{C0C0C0}100 & \cellcolor[HTML]{C0C0C0}100 \\ \hline
                                                                                  & MNIST                                          & 784-200-50-20-10                              & Tanh                            & SGD                             & 0.1                           & 0.25                          & 100                         & 100                         \\
                                                                                  & \cellcolor[HTML]{EFEFEF}Malware                & \cellcolor[HTML]{EFEFEF}3738-3000-100-20-2   & \cellcolor[HTML]{EFEFEF}Relu    & \cellcolor[HTML]{EFEFEF}SGD & \cellcolor[HTML]{EFEFEF}0.1 & \cellcolor[HTML]{EFEFEF}0.25 & \cellcolor[HTML]{EFEFEF}100 & \cellcolor[HTML]{EFEFEF}20  \\
\multirow{-3}{*}{\begin{tabular}[c]{@{}c@{}}Distillation \\ (T=20)\end{tabular}}  & \cellcolor[HTML]{C0C0C0}IMDB                   & \cellcolor[HTML]{C0C0C0}600-100-100-50-2      & \cellcolor[HTML]{C0C0C0}Sigmoid & \cellcolor[HTML]{C0C0C0}SGD     & \cellcolor[HTML]{C0C0C0}0.1   & \cellcolor[HTML]{C0C0C0}0.2   & \cellcolor[HTML]{C0C0C0}100 & \cellcolor[HTML]{C0C0C0}50  \\ \hline
                                                                                  & MNIST                                          & 200-200-100-10                                & Relu                            & Adam                            & 0.001                         & 0.5                           & 100                         & 50                          \\
                                                                                  & \cellcolor[HTML]{EFEFEF}Malware                & \cellcolor[HTML]{EFEFEF}1000-500-200-100-2    & \cellcolor[HTML]{EFEFEF}Relu    & \cellcolor[HTML]{EFEFEF}Adam & \cellcolor[HTML]{EFEFEF}0.001 & \cellcolor[HTML]{EFEFEF}0.5 & \cellcolor[HTML]{EFEFEF}100 & \cellcolor[HTML]{EFEFEF}50  \\
\multirow{-3}{*}{LLE-DNN}                                                             & \cellcolor[HTML]{C0C0C0}IMDB                   & \cellcolor[HTML]{C0C0C0}500-300-200-100-2     & \cellcolor[HTML]{C0C0C0}Tanh    & \cellcolor[HTML]{C0C0C0}Adam    & \cellcolor[HTML]{C0C0C0}0.001 & \cellcolor[HTML]{C0C0C0}0.5   & \cellcolor[HTML]{C0C0C0}100 & \cellcolor[HTML]{C0C0C0}100 \\ 
\hline
\end{tabular}
\caption{\textbf{The hyperparameters of all the investigated DNN models.}}
\label{tab:hyperparameters}
\end{table*}

\section{Evaluation}
\label{sec:eval}

As is described in Section~\ref{sec:intro}, adversarial training~\cite{Goodfellow14} and defensive distillation~\cite{papernot2015distillation} are the most representative techniques that have been proposed to defend against adversarial samples. In this work, we use the proposed approach to train our own adversary resistant DNN (LLE-DNN), and then compare it with those enhanced by these two approaches. 

\subsection{Dataset}
We evaluate our adversary-resistant DNN model by performing multiple experiments on several widely used datasets, including a dataset for malware detection~\cite{berlin2015malicious}, the MNIST dataset for image recognition~\cite{lecun1998mnist} and the IMDB dataset for sentiment analysis~\cite{maas2011learning}. 

\noindent{\textbf{Malware dataset:}}
It is a collection of window audit logs, each of which ties to either a benign or malicious software sample. The dimensionality of the feature-space for each sample is reduced to 10,000 based on the feature selection metric in~\cite{berlin2015malicious}. Each feature indicates the occurrence of either a single filesystem access or a sequence of access events, thus taking on the value of 0 or 1. Figure~\ref{tab:malware} illustrates a subset of features of a software sample. Here, 0 indicates that the sequence of events did not occur while 1 indicates the opposite. For each software sample, it has been labeled with either 1 or 0, indicating malicious and benign software, respectively. The dataset is split into 26,078 training examples, with 14,399 benign and 11,679 malicious software samples,  and 6,000 testing samples, with benign and malicious software samples evenly divided. 

\noindent{\textbf{MNIST dataset:}} 
It is a large database of handwritten digits that is commonly used for training various image processing systems. It is composed of 70,000 greyscale images (of 28$\times$28, or 784, pixels) of handwritten digits, split into a training set of 60,000 samples and a testing set of 10,000 samples. 

\noindent{\textbf{IMDB dataset:}} 
It consists of 25,000 movie reviews, with one half labeled as ``positive'' and the other ``negative'', indicating the sentiment of these reviews. We randomly split the dataset with 70\% movie reviews for training and the remaining for testing. Following the procedure introduced in~\cite{mikolov2013distributed}, we encoded the words in each movie review using a dictionary carrying 5,000 words most frequently used. Then, we utilized a word embedding technique~\cite{mikolov2013distributed} to convert each word into a vector with a dimensionality of 600. For each movie review, we linearly combined the vectors indicating the words appearing in that review, and then treat the embedding as the representation of that movie review. 

\begin{figure*}[t!]
    \centering
    \begin{subfigure}[t]{0.5\textwidth}
        \centering
        \includegraphics[width=0.98\textwidth]{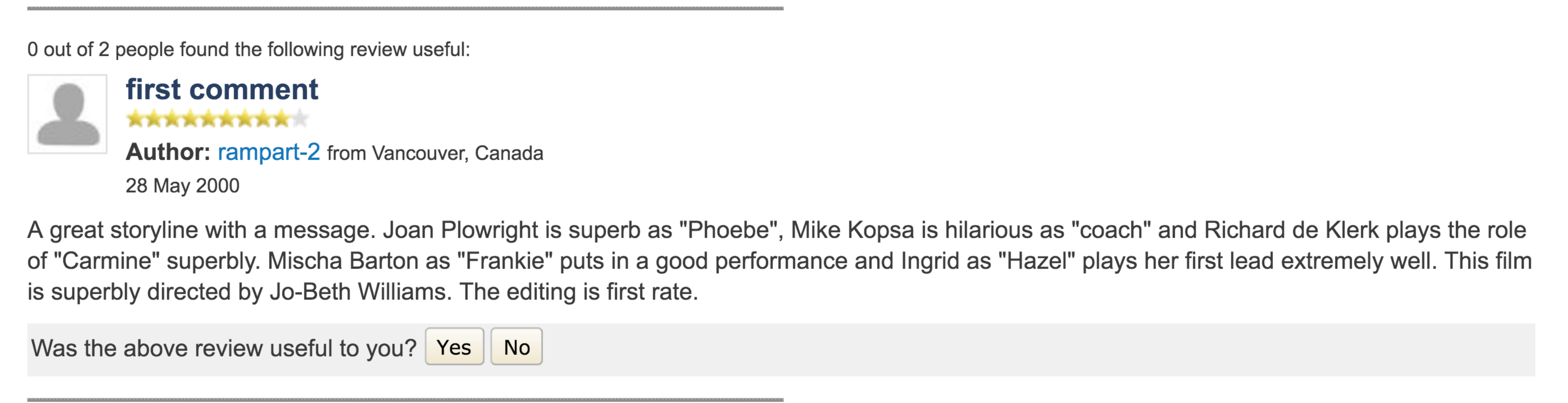}
        \caption{Legitimate movie review.}
    \end{subfigure}%
    ~ 
    \begin{subfigure}[t]{0.5\textwidth}
        \centering
        \includegraphics[width=0.98\textwidth]{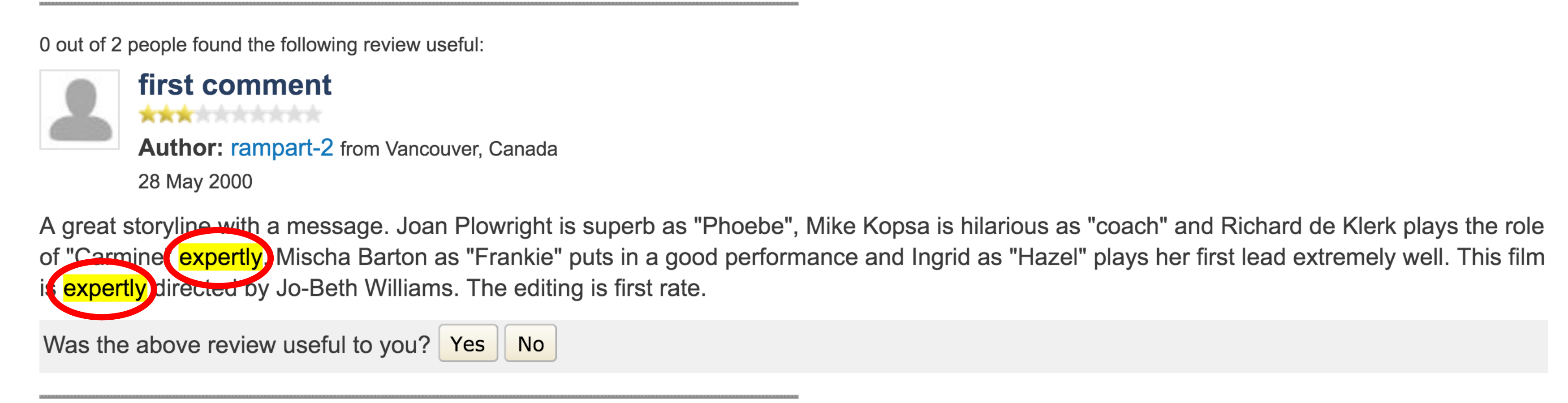}
        \caption{Adversarial movie review.}
    \end{subfigure}
    \caption{A legitimate movie review and its adversarial sample. Note that the replacement words are highlighted.}
    \label{fig:review}
\end{figure*}

\subsection{Experimental Design}

For each application described above, we train 4 DNN models using the traditional deep learning training method, adversarial training~\cite{Goodfellow14}, defensive distillation~\cite{papernot2015distillation} and our own approach. We specify the hyperparameters of these DNNs in Table~\ref{tab:hyperparameters}. We measure their classification accuracy by applying the models to the corresponding testing datasets. By comparing their classification performance, we evaluate the influence that our proposed approach brings to a DNN. More specifically, we examine if LLE-DNN exhibits similar -- if not the same or better -- classification accuracy. 

Since the goal of this work is to improve the robustness of a DNN model, we also evaluate our DNN models' resistance to adversarial samples. In particular, we derive adversarial samples from the aforementioned testing datasets, test them against our DNN model and compare its model resistance with those of DNNs enhanced by the other two techniques~\cite{papernot2015distillation, Goodfellow14}. 

As is discussed in Section~\ref{sec:advlearn}, an attacker crafts adversarial samples through auxiliary models. In Table~\ref{tab:resistance}, black-box and white-box indicate the auxiliary models trained through different schemes. More specifically, black-box represents the auxiliary model trained through the standard deep learning training scheme, indicating an attacker does not have sufficient knowledge about the underlying training algorithm and he can use only a standard approach to train a cross model and craft adversarial samples. White-box represents the auxiliary model trained exactly through the learning schemes proposed as a defense. This simulates a situation where a defense mechanism is publicly disclosed and an attacker exploits that mechanism to produce a highly similar -- if not the same -- model to craft adversarial samples. Note that, for both black-box and white-box tactics, we use the same hyperparameters and training dataset to build auxiliary models. More specifically, our auxiliary model training shares the same hyperparameters and training dataset with the standard DNN shown in Table~\ref{tab:hyperparameters}.

In addition to the methods described in Section~\ref{sec:advlearn}, the crafting of adversarial samples must ensure a slight perturbation introduced to a data sample does not undermine its semantic. In other words, we must make sure that, while misleading a classifier to output the wrong class with high confidence, the perturbation to an image should be nearly indistinguishable to the human eyes, that to a malicious software sample should not jeopardize software functionality nor break its malevolence, and that to a movie review should not break its semantic meaning. In the following, we describe how we fine-tune adversarial samples to preserve semantic for different applications.

\noindent{\bf Malware classification.}
Recall that our malware samples are represented by features, the value of which are binary, indicating the occurrence of an filesystem access or a sequence of access events. When generating adversarial samples, we cannot simply disable filesystem access events in that this might jeopardize the functionality of the software sample and even break down its malevolence. With this in mind, a bit of care must be taken. 

In this work, our experiment follows the approach introduced in Section~\ref{sec:advlearn}. To be specific, we craft adversarial software samples by solving optimization problem~\eqref{eq:adv1} with the setting of zero norm (i.e., $l_0$). This indicates the manipulation to a sample is restricted to flipping binary feature values. In addition, this implies the strongest attack scenario in that optimization problem~\eqref{eq:adv1} carries less constraints making the adversarial samples chosen for our evaluation more impactful. Going beyond adversarial sample crafting approach discussed in Section~\ref{sec:advlearn}, we also restrict that the value change of a feature can be only from 0 to 1 but not the opposite. This amounts to allowing the addition of new filesystem access events only. This manipulation strategy is reasonable for the reason that malware mutation techniques (e.g.,~\cite{MohanH12}) can morph a malware sample by stitching together instructions from benign programs, making the malware perform additional filesystem accesses but not undermining its maliciousness nor its functionality. Since malware manipulation is done with the intent of fooling a malware classifier driven by a DNN, it should be noticed that we do not morph a benign software sample, making it malicious.  

\noindent{\bf Image recognition.}
Image data samples contain less strict semantic than the malware data samples above. To preserve image semantic -- making a perturbation nearly indistinguishable -- we follow the approaches introduced in~\cite{Carlinioakland2017, 42503, Goodfellow14}. More specifically, we selected $l_{0}$, $l_{2}$ and $l_{\infty}$ distance to represent the dissimilarity between an image and its corresponding adversarial sample. Especially, we restrict the $l_{\infty}$ distance in a relative small range (i.e., $\epsilon \leq 0.15$) when crafting adversarial samples.

\noindent{\bf Sentiment Analysis.}
To generate adversarial samples for movie reviews, we again followed the approach introduced in Section~\ref{sec:advlearn}. To be specific, we solved optimization problem~\eqref{eq:adv1} and configured $p$-norm with the setting to $l_2$ and $l_{\infty}$. This is due to the fact that each review is encoded in a vector in which each element is a decimal, and $l_2$ and $l_{\infty}$ distances represent the best measure for the dissimilarity between a movie review and its corresponding adversarial sample.

As is mentioned above, each vocabulary has been encoded in a vector with a dimensionality of 600, and we embedded a movie review by linearly combining corresponding vectors. When generating an adversarial sample by introducing a slight perturbation to the embedding, we cast the perturbation to only one vector. This ensures that we introduce only one word change to that review with the hope that it preserves the semantic meaning of that review as much as possible. However, one word change does not guarantee the invariance of the semantic meaning. For example, it would be obvious alteration to the semantic meaning if the replacement happens to be the negative word in {\em ``... makes it the biggest disappointment I've experienced from cinema in years ..."}. As such, we manually choose the word that incurs minimal semantic change to that movie review. Figure~\ref{fig:review} illustrates a movie review sample and its corresponding adversarial sample generated through this approach.

\subsection{Experimental Setup and Results}

On the datasets described above, we first measure the accuracy of all the aforementioned defense techniques. We then measure their resistance to the adversarial samples crafted through the aforementioned tactics.

\begin{table}[]
\centering
\begin{tabular}{cccc}
\hline
\multirow{2}{*}{Learning technology} & \multicolumn{3}{c}{Accuracy}                                                       \\ \cline{2-4} 
                                     & \multicolumn{1}{l}{MNIST} & \multicolumn{1}{l}{MALWARE} & \multicolumn{1}{l}{IMDB} \\ \hline
Standard DNN                         & 98.45\%                   & 92.97\%                     & 87.89\%                  \\ \hline
Distillation                         & 98.46\%                   & 92.45\%                     & 87.36\%                  \\ \hline
Adv. training                        & 98.77\%                   & 91.48\%                     & 87.67\%                  \\ \hline
LLE-DNN                              & 98.19\%                   & 93.56\%                     & 87.79\%                  \\ \hline
\end{tabular}
\caption{The comparison of classification accuracy on different datasets.}
\label{tab:accuracy}
\end{table}

\begin{figure*}[t]
    \centering
    \begin{subfigure}[t]{0.32\textwidth}
        \includegraphics[width=1.0\textwidth]{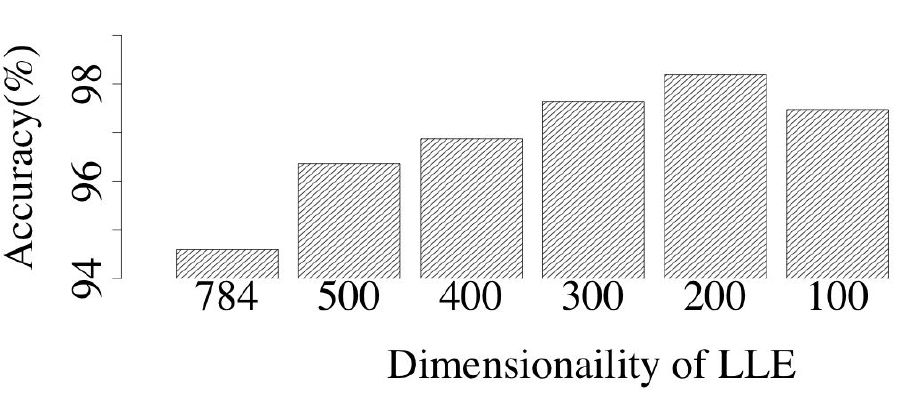}
        \caption{MNIST.}
        \label{subfig:mnist_lle_dim}
    \end{subfigure}
    \begin{subfigure}[t]{0.32\textwidth}
        \includegraphics[width=1.0\textwidth]{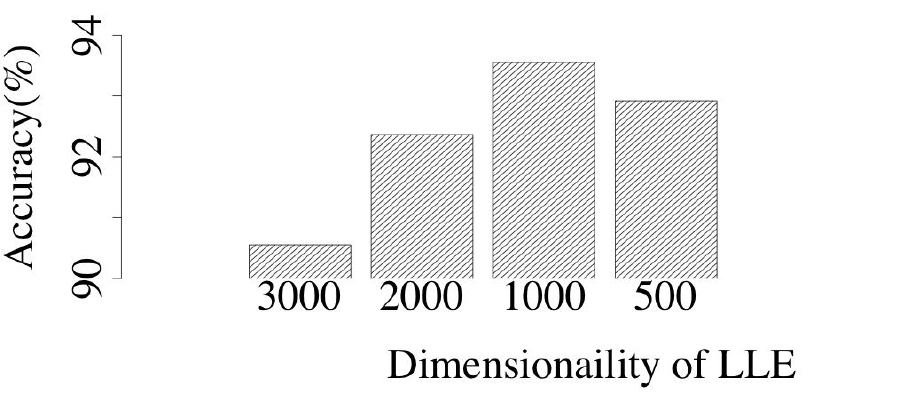}
        \caption{Malware.}
        \label{subfig:malware_lle_dim}
    \end{subfigure}
    \begin{subfigure}[t]{0.32\textwidth}
        \includegraphics[width=1.0\textwidth]{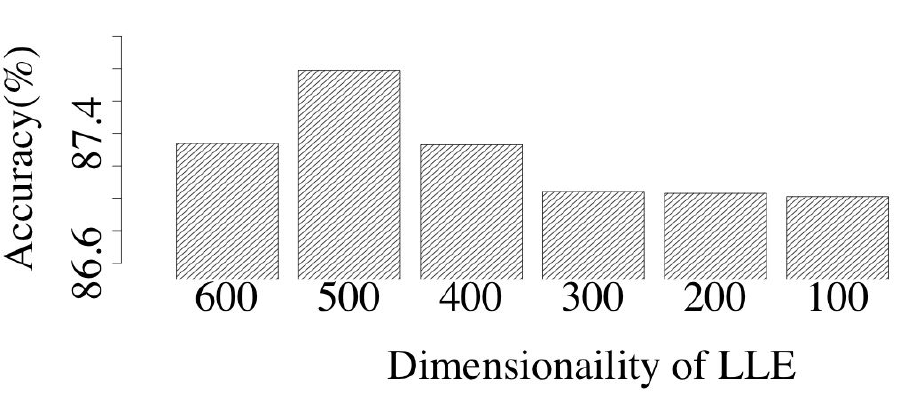}
        \caption{IMDB.}
        \label{subfig:imdb_lle_dim}
    \end{subfigure}
    \caption{\textbf{The variation of classification accuracy vs. the dimensionality of data mappings.}}
    \label{fig:img_lle_dim}
\end{figure*}

\subsubsection{Classification Accuracy}

To identify the optimal dimensionality to which LLE needs to map original data samples, we implemented several LLE-DNNs with different settings of dimensionality of LLE mappings. Figure~\ref{fig:img_lle_dim} shows the impact of dimensionality mapping upon the classification accuracy obtained by LLE-DNN. Across all three datasets, it is easy to observe that, the classification accuracy first increases when the dimensionality of the LLE mappings rises and then starts to decrease. In our experiment, we choose the highest classification accuracy to represent the performance of our LLE-DNN.   

Table~\ref{tab:accuracy} presents the classification accuracy results obtained from all investigated DNNs with respect to the testing datasets. Note that, while prior works (e.g,~\cite{abs-1003-0358, deep-convex-network}) have already demonstrated they can train a DNN with an error rate less than 1\% on the MNIST benchmark, their performance improvement does not result from a DNN but model ensemble or elastic distortions added to training data. To study the influence of our proposed approach upon a standard DNN, therefore, we did not combine models nor augment with artificially distorted versions of the original training samples. The classification accuracy shown in the table has already represented the best performance that a standard DNN can achieve. 

Similar to adversarial training and defensive distillation, the LLE-DNN is quite effective in preserving classification accuracy. This implies our proposed approach well preserves data sample distribution. For the malware classification task, it can be observed that LLE-DNN appears to be better at feature learning, achieving the highest classification accuracy among DNNs that we investigated. This is presumably due to the fact that malware data samples are highly sparse carrying a large amount of redundant information, and the data transformation module in LLE-DNN eliminates those redundancy and ameliorates the learning ability of a DNN.

\subsubsection{Model Resistance}

Table~\ref{tab:resistance} illustrates the DNNs that we investigated as well as their accuracy in classifying adversarial samples. It can be observed that, black-box adversarial samples can cut down the accuracy of the standard DNN to 6.86$\%$, 6.40$\%$ and 7.50$\%$ under the attacks of $l_{\infty}$, $l_{2}$ and $l_{0}$, respectively. In contrast, all of the defense mechanisms investigated demonstrate strong resistance to these black-box adversarial samples. This indicates, without sufficient knowledge on the underlying defense mechanisms, it is difficult for an attacker to craft impactful adversarial samples. In other words, existing defense mechanisms can significantly escalate a DNN's resistance to adversarial samples if one can obscure the design of the defenses.

Despite the improvement in model robustness, we also observe that our LLE-DNN generally exhibits the best resistance to black-box adversarial samples, whereas the defensive distillation approach typically yields the worst resistance. This is presumably due to the fact that, the dimensionality reduction resided in LLE-DNN transforms adversarial samples into a subspace in which they no longer act as outliers, while defensive distillation smooths only a classification decision boundary which does not significantly reduce the subspace of adversarial samples.  

With regard to the white-box setting, we discover both adversarial training and defensive distillation suffer from white-box adversarial samples. Their resistance to white-box adversarial samples is significantly worse than those created under the black-box setting. This observation is consistent with that reported in~\cite{Carlinioakland2017}. The reason behind this is that, both techniques stash away the adversarial sample subspace, but the disclosure of defense mechanisms uncovers the path of finding that subspace.

Different from adversarial training and defensive distillation, our LLE-DNN is naturally resistant to white-box adversarial samples. As is discussed in Section~\ref{sec:framework}, our proposed approach stashes away the adversarial sample subspace and at the same time restricts derivative calculation. Even if our defense mechanism is revealed, therefore, it is still computationally difficult to find adversarial samples. 

To perform quantitative comparison with the other two approaches, however, we approximate the data transformation in the LLE-DNN -- non-parametric dimensionality reduction component -- using a parametric model. To be specific, we choose a DNN to approximate LLE in that a DNN has a large amount of parameters which is typically viewed as the best approximation for non-parametric learning models~\cite{hornik1991approximation}. With the support from this approximation, we treated the LLE-DNN as a white box and generated adversarial samples accordingly. We show its model resistance in Table~\ref{tab:resistance}. It can be observed that, our LLE-DNN still demonstrates strong resistance to white-box adversarial samples even if we substituted LLE to its best approximation. This implies that, there might a theoretical lower bound between a non-parametric model and its parametric approximation, which can naturally serve as a defense against white-box adversarial samples. 

\section{Conclusion}
\label{sec:conclusion}

A Deep Neural Network is vulnerable to adversarial samples. Existing defense mechanisms improve a DNN model's resistance to adversarial samples by using the tactic of security through obscurity. Once the design of the defense is disclosed, therefore, the robustness they provide wane. Motivated by this, this work introduces a new approach to escalate the robustness of a DNN model. In particular, it integrates to a DNN model LLE, a non-parametric dimensionality reduction method. With this approach, we show that one can develop a DNN model resistant to adversarial samples even if he or she reveals its design details (i.e., the underlying training algorithm). By demonstrating the DNNs enhanced by our proposed technique across various applications, we argue the proposed approach introduces nearly no degradation in classification performance. In contrast, for some applications, it even exhibits performance improvement. As part of the future work, we will further explore the performance of this approach in a wider variety of applications across different deep neural architectures.

\bibliographystyle{abbrv}
\bibliography{ref}
\end{document}